\pdfoutput=1

\documentclass{ecai} 
\newcommand{\ignore}[1]{}

\usepackage[colorinlistoftodos]{todonotes}

\usepackage{times}
\usepackage{latexsym}
\usepackage{booktabs}
\usepackage{multirow}
\usepackage{pdflscape}
\usepackage{rotating}
\usepackage{float}
\usepackage{cancel}
\usepackage{xcolor}
\usepackage{amsmath}
\usepackage{hyperref}
\usepackage[T1]{fontenc}

\usepackage[utf8]{inputenc}

\usepackage{microtype}

\usepackage{inconsolata}

\usepackage{graphicx}
\usepackage{amssymb}

\newcommand{\BibTeX}{B\kern-.05em{\sc i\kern-.025em b}\kern-.08em\TeX}
\newcommand{\setR}{\mathbb{R}}

\begin{document}
\begin{frontmatter}


\paperid{1202}

\title{Prediction is not Explanation:\\Revisiting the Explanatory Capacity of Mapping Embeddings}

\author{\fnms{Hanna}~\snm{Herasimchyk}}
\author{\fnms{Alhassan}~\snm{Abdelhalim }}
\author{\fnms{Sören}~\snm{Laue}} 
\author{\fnms{Michaela}~\snm{Regneri}\thanks{Corresponding Author. Email: \url{michaela.regneri@uni-hamburg.de}}  }


\address{Universität Hamburg, Hamburg, Germany}


\begin{abstract}
Understanding what knowledge is implicitly encoded in deep learning models is essential for improving the interpretability of AI systems. This paper examines common methods to explain the knowledge encoded in word embeddings, which are core elements of large language models (LLMs). These methods typically involve mapping embeddings onto collections of human-interpretable semantic features, known as feature norms. Prior work assumes that accurately predicting these semantic features from the word embeddings implies that the embeddings contain the corresponding knowledge. We challenge this assumption by demonstrating that prediction accuracy alone does not reliably indicate genuine feature-based interpretability.

We show that these methods can successfully predict even random information, concluding that the results are predominantly determined by an algorithmic upper bound rather than meaningful semantic representation in the word embeddings. Consequently, comparisons between datasets based solely on prediction performance do not reliably indicate which dataset is better captured by the word embeddings. 
Our analysis illustrates that such mappings primarily reflect geometric similarity within vector spaces rather than indicating the genuine emergence of semantic properties.
\end{abstract}

\ignore{Understanding what knowledge is encoded in neural network representations is essential for improving the interpretability of AI models. This paper examines common methods used to explain the knowledge encoded in word embeddings used in current large language models (LLMs.) These methods typically involve mapping word embeddings onto human-interpretable semantic features, known as feature norms. Prior work assumes that accurately predicting these semantic features from the word embeddings implies that the embeddings contain the corresponding knowledge. We challenge this assumption by demonstrating that prediction accuracy alone does not reliably indicate true property-based interpretability.

Through controlled experiments, we show that these methods can successfully predict even random information, highlighting that results are predominantly determined by an algorithmic upper bound rather than meaningful semantic representation in the word embeddings. 
By systematically contrasting categorical and continuous feature norms and using appropriate evaluation metrics, we clarify the limitations and potential of these mapping methods used for explainability. Our analysis illustrates that such mappings primarily reflect structural equivalences within vector spaces rather than indicating the emergence of semantic properties.}

\end{frontmatter}

\section{Introduction}
Large Language Models (LLMs) have emerged as a core tool and central research area in natural language processing. While their capabilities are impressive, we still need to understand how they gain such excellent performance apart from using vast amounts of data and parameters.  While many algorithms exist for explaining LLMs and their individual components, many of those approaches are complex and not fully understood themselves. Differentiating faithful explanations from misleading correlations is hard, and "Clever Hans" effects \citep{shapira-etal-2024-clever} also occur for explainability methods.

In this paper, we focus on explanation methods for word embeddings, which serve as the input layer for LLMs. A popular approach for explaining which kind of knowledge is implicitly learned when training such embeddings on large amounts of text is called "property inference"~\cite{Rosenfeld_Erk_2023}. Property inference aims to detect knowledge (the "properties") about a word implicitly stored in its embedding vector by predicting ("inferring") the properties in question from the embedding. 
To achieve this, a predictive model is trained to map an embedding vector onto a corresponding set of properties, often taken from curated datasets known as feature norms. These norms provide interpretable, human-annotated semantic features that vary in their relationship to their associated words. 
Several studies used such norms and property inference to demonstrate that emergent knowledge in embeddings contains semantic, perceptual, or neurological features~\citep{abnar-etal-2018-experiential,chersoni-etal-2021-decoding,chronis-etal-2023-method,fagarasan-etal-2015-distributional,hollenstein-etal-2019-cognival,nair-etal-2020-contextualized}. 

The common assumption underlying these studies is that if a model can accurately predict semantic properties from an embedding, then the embedding must encode that knowledge. However, this assumption has not been rigorously validated, and the explanatory capacity of such methods remains uncertain.

In this work, we want to validate and differentiate the explanatory potential of property inference methods. To do so, we apply two methods commonly used in previous studies (Partial Least Squares Regression and Feed Forward Neural Networks) to map word embeddings from BERT \cite{devlin-etal-2019-bert} to three different feature norms (two categorical norms and one with continuous features). We extend previous studies by providing multiple ablation experiments and upper bounds, to put the resulting numbers into perspective. Our results show that while information overlap between feature norms and embeddings is predictable, the results are not faithful explanations of emergent properties, but rather a different type of geometric similarity.
Significant predictive performance can even be achieved for nonsensical or random features, calling into question whether such methods faithfully explain latent knowledge in embeddings.

Our main contributions are as follows:
\begin{itemize}
\item We show that common property inference methods do \emph{not} provide faithful explainability. While high information overlap does lead to good results, good results do not indicate high information overlap. 
\item We provide a detailed set of experiments to assess the explanatory potential of property inference methods.
\item We show that mapping embeddings measures geometric similarity, and we distinguish this from explaining property knowledge.
\end{itemize}
After revisiting previous work (Sec.~\ref{background}), we introduce the methods we use (Sec.~\ref{methods}) and our experimental setup (Sec.~\ref{experiment}). In several ablation experiments, we show that the methods have a low upper bound, which determines more of the results than the information overlap between embeddings and norms (Sec.~\ref{challenge}). Our last argument shows that the prediction methods explain geometric similarity but not individual properties (Sec.~\ref{exp:structure}).
we conclude with a discussion of the results (Sec.~\ref{discussion}), followed by a summary and ideas for future work (Sec.~\ref{conclusion}). 
Code and data \cite{PredictionIsNotExplanation:Code} as well as additional evaluation results 
(in the Appendix)
are provided as supplementary material.\section{Background and related work}\label{background}

Research on embedding explainability was done long before LLMs existed, starting with the first distributional models that derived word meaning from co-occurrence frequencies~\citep{distributional:first}. Originally, those models had directly interpretable dimensions because they denoted context words as features. More advanced models like Word2Vec~\citep{Mikolov2013EfficientEO} or GloVe \cite{pennington-etal-2014-glove} used occurrence probabilities rather than frequencies and condensed the vector spaces to fewer dimensions, so the resulting embedding dimensions were not directly interpretable anymore. Embeddings of this type excelled in Natural Language Processing tasks like word sense discrimination \citep{schutze-1998-automatic} or text classification \citep{glavas-etal-2019-semantic}. At the same time, such embeddings have also been applied to tasks related to cognitive science, like text comprehension \citep{Foltz01011998}.

Sets of cognitively plausible features were originally considered an alternative to distributional embeddings~\citep{10.3115/981623.981633} but now serve as a gold standard for potentially emergent knowledge in generative models. Property inference is one such approach for interpretation that maps curated sets of interpretable properties to embeddings. This explainability method has been used to decode semantic features~\citep{chersoni-etal-2021-decoding,chronis-etal-2023-method,nair-etal-2020-contextualized}, perceptual associations \citep{fagarasan-etal-2015-distributional}, and even neurological patterns from fMRI and EEG data \citep{abnar-etal-2018-experiential,hollenstein-etal-2019-cognival}. Most often, partial least squares regression (PLSR)~\citep{10.1007/11752790_2} or feedforward neural networks (FFNNs) are used for the mapping, but other approaches have been tested (\citet{Rosenfeld_Erk_2023} provide an overview).
As a linear mapping method, PLSR has also found widespread use in fields such as medicine~\citep{Palermo:2009aa}, chemistry~\citep{WOLD2001109}, or economics~\citep{PLS-Eco} due to its robustness in handling high-dimensional and collinear data. 
A common problem with taking the resulting predictions as causal interpretations is that it is difficult to dissect whether the results are, as expected, due to information overlap, whether they stem from the analyzed information source, or are correlations with other features, and what role methodological artifacts play (see~\citet{lu2023emergent} for an overview of end-to-end explanations for LLMs). 

No extensive analysis has shown the explanatory capacity of mapping approaches for embeddings and language-related properties so far. We bridge this gap by providing a set of control experiments and some background from Machine Learning theory to show which type of knowledge the methods can predict and that they cannot explain property-based knowledge.

\section{Methods} \label{methods}
\emph{Word embeddings} are mathematical representations of words in the form of high-dimensional vectors derived from large amounts of text. They are optimized to represent words that tend to occur in similar contexts as nearby vectors in the vector space, and words that do not share contexts as vectors with a large distance. 
Formally, we denote word embeddings as a matrix $X\in\setR^{n\times d}$ , where $n$ is the number of words or concepts and  $d$ is the embedding dimension.

To detect potentially emergent knowledge, we use so-called \emph{feature norms}. Feature norms are structured, partially human-annotated datasets that map concepts to interpretable features. Concepts represent a specific meaning. They are in a many-to-many-relationship with words: An ambiguous word can denote different concepts (like "bank" is either related to money or a river), and a concept can have multiple words as realizers (the person instructing people at a gym can be a "coach" or a "trainer"). In the following, we will refer to \emph{concepts} rather than \emph{words}.\footnote {Note that for very few concepts, ambiguity is a challenge in the mapping approaches, because different meanings of "bank" are a) marked in the norms and b) will be mapped to the same vector in the embeddings. This does not affect our results because this only concerns very few concepts and only the McRae norms. We will use the same embedding vector for concepts named with the same word.}
The features associated with the concepts can be either qualitative from categorical distributions (e.g., “ravens” have the feature “bird”) or quantitative ratings of feature association strength (e.g., the feature “bright” for the concept "sun" has an association strength of 100\%).
We represent feature norms as a matrix $Y\in\setR^{n\times m}$, where $m$ is the number of annotated features. For the sake of simplicity, we use the terms \emph{features} and \emph{properties} interchangeably.

\begin{figure}[t]
\centering
\includegraphics[width=\columnwidth]{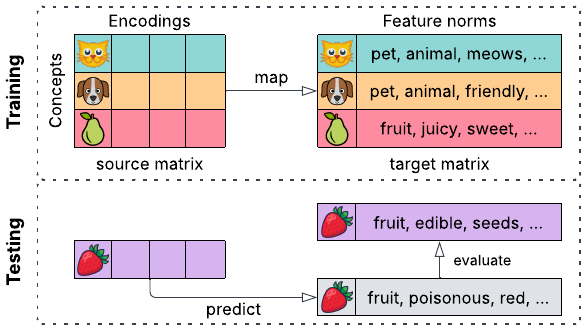}
\caption{The standard pipeline for explaining knowledge in embeddings by mapping them to feature norms.\strut\\\\}
\label{fig:pipeline}

\end{figure}

Using \emph{property inference}, previous work has attempted to bridge word embeddings and feature norms to study which types of interpretable knowledge emerge in embeddings~\cite{Binder:2016aa,chersoni-etal-2021-decoding,chronis-etal-2023-method,Rosenfeld_Erk_2023}. By training a predictive model $f$ to map from word embeddings $X$ to feature norms $Y$, we seek to reveal the implicit knowledge captured by the embeddings $X$, i.e, 
\[
f\colon \setR^{n\times d} \to \setR^{n\times m},\quad X\mapsto Y.
\]
Figure~\ref{fig:pipeline} illustrates this generic experimental setup for knowledge discovery in embeddings via property inference. 

Two widely used methods for this mapping $f$ are \emph{partial least squares regression} and classical \emph{feedforward neural networks}. PLSR is a linear technique that identifies a lower-dimensional latent space in which the covariance between the input embeddings and output features is maximized. Formally, it maps an input matrix $X\in\setR^{n\times d}$ to an output matrix $Y\in\setR^{n\times m}$, by projecting both $X$ and $Y$ into a shared latent space of dimension $k$. FFNNs, in contrast, offer a nonlinear approach. In the context of property inference, the FFNN is typically a fully connected feedforward neural network with $d$  input nodes (matching the embedding dimensionality), $m$ output nodes (matching the number of features), and a single hidden layer of size $k$. The hidden layer uses the tanh activation function. This architecture enables the network to learn flexible mappings from embeddings to features.

PLSR and FFNNs share that they map their input to an intermediate, lower-dimensional latent subspace of dimension $k$. The transformed representation is then used for regression onto the output data. Both methods differ slightly in how they define this lower-dimensional space. PLSR explicitly constrains the latent space to a Krylov subspace \citep{Loefstedt24}, ensuring that the reduced representation captures principal variations (i.e., linear combinations of input features) relevant for predicting the output. In contrast, FFNNs learn their latent representation through optimization, where the training method (e.g., SGD or Adam), along with its learning rate and other hyperparameters, implicitly acts as an additional regularization mechanism to control model complexity~\citep{SmithDBD21}.

Both methods are also tightly connected to the method of reduced-rank regression~\citep{IZENMAN1975248}. Reduced-rank regression is a linear mapping, where the coefficient matrix is constrained to have a rank of at most $k$. If the activation function in the FFNN is replaced by a linear activation function, reduced-rank regression and FFNN coincide. Hence, PLSR and FFNN, the two most commonly used methods for mapping word embeddings onto feature norms, are mathematically very similar. Their close connection and only minor difference explain why their empirical performance typically differs only slightly in practice. We will also verify this in our experiments.

\section{Data and experimental setup} \label{experiment}

The following describes a typical experimental setup for mapping distributional embeddings to feature norms to infer property knowledge. First, we describe the norms we use and then several evaluation metrics from the literature. We then detail the search for an optimal hyperparameter $k$, showing that many previous approaches tend to overfit by choosing $k$ too high. Finally, we report the results, showing that with an optimal $k$, PLSR and FFNN do not differ in performance.
\subsection{Feature Norms}

Following previous work, we use three feature norms for our experiments:
\begin{itemize}
\item \textbf{McRae:} The McRae norms \citep{featurenorms_2005} contain 541 concrete nouns as concepts and 2526 (categorical) features, amounting to 7259 concept-feature pairs. Each concept has between 6 and 26 features, with an average of 13.4 features per concept. The McRae norms come with well-curated meta-annotations, indicating the semantic relations of concepts and associated features (e.g., "bird" and "raven" are in a \emph{taxonomic} relationship, with the feature "bird" being the \emph{superordinate}).
\item \textbf{Buchanan:} The crowdsourced norms by \citet{buchanan_norms} contain 4436 concepts (70\% of nouns, 15\% adjectives, 12\%~verbs, and an 
\emph{other} category) and 3981 unique features, with 69238 concept-feature-pairs in total. Each concept has between 5 and 148 features, with 15.6 features per concept on average.
\item  \textbf{Binder:} The Binder dataset \citep{Binder:2016aa} contains continuous features for 535 concepts with one duplicate (which we remove). Each concept has aggregated crowdsourced ratings for 65 dimensions (like \emph{taste} or \emph{bright}), indicating how strongly related concepts are to the respective feature dimension. In rare cases, the mapping was not applicable (e.g., \emph{caused} for adjectives), and the norm states “na”. Following previous work \citep{chronis-etal-2023-method}, we remove the three features that contain any “na” entries. This amounts to 62 dimensions and 33108 valid averaged feature pair ratings between 0 and 6. 
\end{itemize}
To obtain a processable representation from the categorical norms (McRae and Buchanan), the concept-feature pairs are translated to a matrix $ Y \in \mathbf{R} ^ {n \times m}$ such that each concept vector has a positive value for a feature dimension if the feature representing the dimension is present for the concept, and zero otherwise. For example, the McRae norm is translated into a matrix $ Y \in \mathbf{R} ^ {541 \times 2526}$. In line with \cite{chronis-etal-2023-method}, we use the "production frequency" (the number of subjects out of a possible 30 who listed that feature for that concept in the norms) as a feature value if the feature is present. The Buchanan norm is translated into a matrix $ Y \in \mathbf{R} ^ {4437 \times 11443} $. Following \cite{chronis-etal-2023-method}, we use the "Normalized Translated" column that contains the frequency of the lemmatized features divided by the sample size as a feature value if the feature is present. The sample size varies from 20 to 177, with a median of 30.

Note that this results in very sparse matrices for the two categorical norms (McRae and Buchanan) because they have a large number of dimensions, most of which are zeros. The Binder norm is a dense matrix containing only 62 dimensions but no zero values.
In our experiments, we will show that the different natures of those three norms lead to different pitfalls for using property inference methods. The sparsity of McRae and Buchanan induces very low upper bounds for the prediction, while the dense Binder norms can yield unexpectedly good evaluation results for nonsensical feature values.

\subsection{Evaluation measures}
Different measures are used in previous work:
\begin{itemize}
\item \textbf{Mean Squared Error (MSE)} indicates how much the predicted numerical values differ from the test data. This measure indicates how well the model fits the data and is heavily influenced by feature scaling. For rating correctly predicted properties, MSE is unsuitable because properties can be correctly predicted with a large error simply by ranking them correctly. What MSE indicates is the maximal capacity of a model to fit the data and thus can identify optimal model parameters.
\item \textbf{Correlation coefficients:} Correlation (mostly Spearman's $\rho$) is used for mapping continuous norms like Binder because they were designed to compare feature rankings rather than indicate property presence or absence. For sparse feature norms like McRae and Buchanan, the ranking-dependent correlation values are unsuitable because most entries are zeros and, thus, are not ranked at all. 
\item \textbf{Precision, Recall, F1:} The standard measures assess the proportion of correctly predicted features for each concept. One way to use them is to judge the top n features (Precision@N, Recall@N, F1@N) returned by the model (corresponding to the n dimension with the highest numerical values) and compare them with the feature base from the norm. For McRae and Buchanan, this indicates whether the highest-ranked features are features associated with the respective concept. This measure is not applicable to continuous norms because each feature has a positive value. A variant is to take the Average Precision / Recall / F1 @N, thus averaging over Precision@1, Precision@2, ... until Precision@N. This implicitly gives more weight to erroneous features with higher rank. Another flavor is to set N individually for each concept to the number of features given for this concept, which assures that 100\% is possible but makes the individual concepts incomparable.
Because those measures are all correlated, we only report F1@N as a generic measure for categorical norms.
\item \textbf{Neighborhood Accuracy@N}: Rather than feature retrieval, neighborhood accuracy measures how accurately the feature space of the target matrix is reproduced in terms of concept similarity. For computation, the top N nearest neighbors (according to cosine distance) of the predicted vector are compared to the gold standard.
\end{itemize}
%
%
%

We will report F1@10 (F1) for McRae and Buchanan, and Spearman ($\rho$) for Binder.

\subsection{Model setup} \label{sec:method}

As a starting point, we reproduced the experiments from \citet{chronis-etal-2023-method} and \citet{chersoni-etal-2021-decoding} to explain knowledge in BERT embeddings by mapping the embeddings to different feature norms.
We use the BERT base model\footnote{\href{https://huggingface.co/google-bert/bert-base-uncased}{huggingface.co/google-bert/bert-base-uncased}} \citep{devlin-etal-2019-bert} to get input embeddings for each concept.
Specifically, we take embeddings from the 0th layer of BERT to construct the input matrix $ X \in \mathbf{R} ^ {n \times d} $ where $n$ is the number of concepts in a feature norms dataset and $d = 768 $ is the number of embeddings dimensions in BERT.  If a concept word consists of multiple tokens, the embeddings of tokens are averaged. Obtaining such embeddings for an individual word is described as "decontextualized" embedding \cite{soper-koenig-2022-polysemy} or type-level embedding \cite{chronis-etal-2023-method}. 

We use PLSR and FFNNs to predict feature norms from BERT embeddings. For PLSR, we use the scikit-learn implementation, setting the maximum number of iterations to 1000. The optimal $k$s we found are 25 for McRae, 80 for Buchanan, and 20 for Binder.
For FFNNs, we use a single hidden layer with $\tanh$ activation function and a hidden layer size equivalent to $k$ for PLSR (25 for McRae, 80 for Buchanan, 20 for Binder). We train the FFNN for 50 epochs with a learning rate of $10^{-4}$, MSE as the loss function, and the Adam optimizer.
We evaluate everything using 10-fold cross-validation.

\subsection{Hyperparameter tuning} \label{exp:hyper}
\begin{figure}
\centering
    \includegraphics[width=\columnwidth]{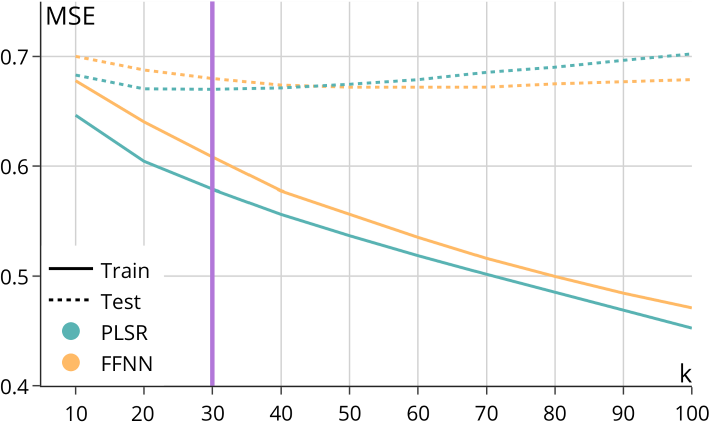}
    
    \caption{MSE of PLSR and FFNNs on training vs. test data (McRae), $k$/hidden layer size on the x-axis. A line marks the optimal parameter.\strut\\\\}
    \label{fig:overfitting}

\end{figure}
Previous approaches \cite{chersoni-etal-2021-decoding,chronis-etal-2023-method} mostly preferred PLSR as a mapping method because it seemed to perform better than FFNNs. We already argued in Sec.~\ref {methods} that both methods are essentially equivalent for property inference. On closer inspection, we see that the differences mainly result from suboptimal hyperparameters. The resulting models exhibit overfitting, i.e., they overadapt to the small amounts of training data and no longer generalize.
The complexity of both methods is primarily governed by the hyperparameter $k$. If $k$ is set too high, PLSR and FFNNs tend to overfit, meaning they memorize the training data but fail to generalize well to unseen test data. 
In previous work, the PLSR parameter $k$ and the FFNN hidden layer size were often set too high, typically ranging from 50 to 300, leading to overfitting. However, both methods perform nearly identically when the optimal parameter settings are used. Fig.~\ref{fig:overfitting} illustrates this equivalence by plotting Mean Squared Error (MSE) for training and test data using PLSR and FFNNs on the McRae dataset. Additional graphs for other feature norms and F1-score results are provided in the technical Appendix.

The onset of overfitting is not always evident from F1-score or Spearman’s $\rho$. While these metrics assess ranking consistency, MSE provides a more direct indicator of overfitting. In some cases, F1-scores remain stable despite overfitting, whereas MSE highlights differences in model adaptation between training and test data. The optimal “elbow” point, where MSE is minimized on test data while avoiding overfitting, is the same for PLSR and FFNNs. 

Given these findings, the choice of mapping method (PLSR vs. FFNN) does not significantly impact results. Instead, dataset constraints and hyperparameter selection are the primary determinants of performance. Different factors in the datasets restrict the generalization capacity: Sparsity (McRae and Buchanan) and a small number of data points (McRae and Binder) limit the predictive capacity. The control experiments we introduce in Sec .~\ref {challenge} will further elaborate on the consequences of those restrictions from data-related properties.

\subsection{Results}\label{exp:eval}

\begin{table}
    \centering
    \caption{Mapping BERT embeddings to different feature norms. Sys = full model, Rand = random baseline.\strut\\}

    \begin{tabular}{l cc cc}
        \toprule
        Norm &  \multicolumn{2}{c}{PLSR} & \multicolumn{2}{c}{FFNN} \\
        \cmidrule(lr){2-3} \cmidrule(lr){4-5}
        & Sys & Rand & Sys & Rand  \\
        \midrule
        McRae (F1) &  0.25 &  0.01 &  0.23 &  0.01 \\
        Buchanan (F1) &  0.18 & 0.01 &  0.17 & 0.01 \\ \midrule
        Binder ($\rho$)& 0.74 & 0.01 &  0.71 &  0.00 \\
        \bottomrule
    \end{tabular}
    \label{tab:startexperiment}
\end{table}

We provide the results for both PLSR and FFNNs for mapping BERT to each of the three feature norms and compare them with a naive random baseline. For the baseline ("Rand"), we fill the feature norm matrix with random numbers and try to predict the correct features from those. All experimental results are shown in Table~\ref{tab:startexperiment}.

The systems perform significantly better than a random baseline. While those numbers are typically used to infer that there is substantial shared knowledge between the norms and the embeddings, the results, as they stand, do not give sufficient evidence for this, which we will empirically show with multiple ablation experiments.

\begin{table*}
\centering
    \caption{Overview of our main findings and the experimental setups that led to them. \emph{Name} are the systems' names used in the evaluation tables, \emph{Source} and \emph{Target} are the matrices mapped to each other. Our own contributions are in boldface. \strut\\}
\vspace*{5mm}
    \begin{tabular}{p{45mm} p{15mm} p{15mm} p{28mm}  p{50mm} }
        \toprule
 \textbf{Result} &  \textbf{Name} & \textbf{Source} & \textbf{Target} & \textbf{Description}  \\
        \midrule
     \emph{Reproducing Experiments from the Literature}, Sec.~\ref{exp:eval} &    Sys & BERT & Original Feature Norms &  Standard experimental setup using PLSR and FFNN \\
        & & & &  \\
          \textbf{PLSR \& FFNN are equivalent}, Sec.~\ref{methods},\ref{exp:hyper},\ref{exp:eval} &    Sys & BERT & Original Feature Norms & Hyperparameter tuning for standard experiments \\
        & & & & \\
      \emph{Completely random features are not predictable}, Sec.~\ref{exp:eval},\ref{sec:shuffle},\ref{exp:structure} &  Rand(om) & BERT & Norms with Completely Random Values &  All values in the Norm Matrix replaced by a random number (sparse norms thus turn into dense norms). Standard baseline. \\     & & & & \\
      
       \textbf{Perfect information overlap leads to low prediction results}, Sec.~\ref{sec:upper}  &     Upper Bound  & Norm Variant & Source Matrix &  Mapping any matrix to itself. \\
       
       & & & & \\

      \textbf{Sparse random features are better predictable than the baseline}, Sec.~\ref{sec:shuffle}  &  Shuffle  & BERT & Norms with random Features &  Assigns random features to each concept. The number of features (and thus the numbers of zeros) is the same as in the original norms.  \\      & & & & \\
        
\textbf{Corrupting essential linguistic features does not yield a significant performance drop.},  Sec.~\ref{sec:tax}  &        Taxonomic Shuffle & BERT & McRae norm with random linguistic hypernyms &Replacing hypernyms with random other hypernyms  \\     & & & & \\
        
  \textbf{Correlation coefficients for dense norms can be high for nonsensical rankings},  Sec.~\ref{sec:cdiff}  &   CDiff  & BERT & Norms with nonsensical Feature Rankings &  Replaces all values by the absolute numerical difference between concept character count and feature character count.  \\   

        \bottomrule
    \end{tabular}
    \label{tab:experiments}
\end{table*}

\section{Measuring explanatory capacity}\label{challenge}
In the following, we will show that algorithmic boundaries largely determine the results we see, and that their influence overwrites the information overlap that should be explained. For the sparse norms, we will show that random features can be predicted to some degree, and that the evaluation results do not allow for separating information overlap from methodological upper bounds as a delimiting factor for performance. For the dense Binder norms, we will show evidence that nonsense scores can lead to equally good results as the actual norm mapping. Given the equivalence of PLSR and FFNN shown in Sec.~\ref{sec:method}, we provide the results of PLSR only and show the results for FFNN in the technical Appendix. Table \ref{tab:experiments} gives an overview over all our experiments with the respective outcome.
\subsection{Upper bounds for property inference}\label{sec:upper}
To put the evaluation results into perspective, we compute an upper bound that indicates the methodological limits for property inference. 
We define an experiment with perfect information overlap, namely mapping a matrix to itself. 
For example, as an upper bound for the mapping of BERT to McRae, we try to predict McRae features with McRae vectors as input, instead of predicting McRae features from BERT vectors. The upper bound for Partial Least Squares Regression for a target matrix $Y$ (and any source matrix $X$) is thus computed as $\text{Upper}(Y) = \text{PLSR}(Y,Y)$.

\begin{table}[H]
\centering

\caption{\strut\\ Upper bounds for property inference with PLSR.\strut\\ }
\vspace*{3mm}
\begin{tabular}{l cc cc cc}
\toprule
\textbf{Norm} & \textbf{Sys} &  \textbf{Upper} &   \textbf{Rand} &  \textbf{Rand-Upper} \\
\midrule
McRae (F1)&  0.25 & 0.27 &  0.01 & 0.01 \\
Buchanan (F1) & 0.18 & 0.22 & 0.01 & 0.01 \\ \midrule
Binder ($\rho$) & 0.74 & 0.90 & 0.01 & 0.52 \\
\bottomrule
\end{tabular}
 \label{tab:upper}
\end{table}

Table \ref{tab:upper} shows the original experiment from Sec.~\ref {exp:eval} (PLSR), along with the upper bounds for the mapping system and the random baseline. The prediction quality is close to the (very low) upper bound for the sparse norms. At the same time, it is evident that the matrix with random numbers cannot be predicted at all, simply due to the missing regularities in the dataset. This is different for the dense Binder norms: To some degree, the random feature rankings are learnable, but cannot be predicted from BERT vectors. In small-scale settings, even random numbers expose some learnable structure. 
With this low maximum performance for McRae and Buchanan, it's questionable how much of the result is restricted by the information overlap, and how much by the method. In particular, comparing two results does not license direct conclusions about differences in information overlap. We will provide further experiments to show that other types of random features can, in fact, be predicted, and that the influence of the upper bound is not separable from the influence of information overlap.

\emph{Takeaway: The upper bound for perfect information overlap with property inference is very low for sparse norms, and prediction accuracy is close to the upper bound. }
\subsection{Sparse norms: predicting random features}\label{sec:shuffle}
Our random baseline ("Rand") fills the complete norm matrix with random numbers. The resulting matrix is thus not sparse anymore. To isolate the influence of the data structure on property predictability, we introduce another baseline that tries to predict random features, but retains the original sparsity of the matrix. The \emph{Shuffle} baseline assigns random features to each concept, but maintains the same number of features as in the gold standard. So for a concept with 4 features, we chose 4 random features out of all possible features in the respective norm and assign them a feature value of 1.\footnote{Note that we used different feature values in the original setup, but for the sparse norms, the absolute value of the features does not have significant influence on the results.}
 \begin{table}[H]
\centering
\caption{Results \& upper bounds for shuffled features and random baseline.\strut\\ }

\begin{tabular}{l cc cc cc}
\toprule
\textbf{Norm}  &  \textbf{Sys} &  \textbf{Upper} &  \textbf{Shuffle} &  \textbf{Shuf-Upper} &  \textbf{Rand}  \\
\midrule
McRae (F1)&  0.25 & 0.27 & 0.10 & 0.13 & 0.01  \\
Buchanan (F1) & 0.18 & 0.22 & 0.06 & 0.11 & 0.01  \\ \midrule
Binder ($\rho$) & 0.74 & 0.90 & 0.30 & 0.59 & 0.01  \\
\bottomrule
\end{tabular}
 \label{tab:shuffle}
\end{table}
Table \ref{tab:shuffle} shows the results. For reference, we also show the completely random baseline. We can see that the random features in \emph{Shuffle} still perform significantly better than the completely random baseline (\emph{Rand}). As we see, for the sparse norms, even random features can be learned to some degree. Further, the upper bound for the random features \emph{Shuf-Upper} is very close to the system's results. So BERT vectors as input perform almost as well for predicting random features as does the random feature matrix itself  (with both results being relatively low).
The results also seem to imply that the random features in the sparse \emph{Shuffle} baseline are better represented in BERT than the random features from the \emph{Rand} baseline, which is a wrong conclusion, because none of the random features should emerge in BERT. This emphasizes that the evaluation of correctly predicted features is, at least, inconclusive about the information overlap between embeddings and norms, simply because the upper bound is so low, and because some degree of randomness can be learned as long as the matrix structure is still sparse enough.

\emph{Takeaway: Random features can be predicted to some degree, and it is not distinguishable whether the result is limited more by (the lack of) information overlap or by the upper bound induced by the dataset's predictability.}

\subsection{Prediction with corrupted semantic features}\label{sec:tax}
In an additional experiment, we want to show that even the random assignment of semantically important features does not affect the results as much as one would assume. For this experiment, we make use of the detailed annotation in the McRae norms (and thus can only show the results on those). Particularly, we target concept-feature pairs in taxonomic relations (hypernyms). Taxonomic relations are \emph{''is-a''} relations (e.g., ``ravens'' are ``birds''), which are essential for any kind of logical reasoning.

We would expect that corrupting the taxonomic concept structure leads to a significant performance drop for property inference. To examine this, we assign each concept random hypernyms and leave all other features unchanged.  E.g., the concept ``raven'' has three hypernyms in the McRae norms (``bird'', ``animal'' and ``scavenger''). In this experiment, we would set the values for those three features to 0, and assign a positive value to a random selection of three other possible hypernyms (e.g., ``fruit'',``musical instrument'', ``building''). We do not control for the compatibility of the random hypernyms among each other (buildings cannot be fruit), so this can yield many impossible constructs.

We then apply property inference as before. The linguistically destructive change hardly changes the results: F1@10 is 0.23 (compared to 0.25 for the original experiment). So, even corrupting linguistic core features, which have been shown to be reflected in embeddings, does not affect the overall results as one would expect. This is particularly surprising because building an obviously wrong set of taxonomic relationships drastically changes the potential information overlap, because it has long been shown that distributional similarity is highly correlated with taxonomic similarity \citep{pado-lapata-2007-dependency}.

\emph{Takeaway: Corrupting essential linguistic knowledge like taxonomic features does not change the prediction results.}

\subsection{Dense norms: predicting random rankings}\label{sec:cdiff}
So far, the performance of property inference on the dense Binder norms in our experiments shows the results one would expect if we actually measured information overlap. The influence of the data-induced upper bounds and random artifacts seems to concern exclusively the sparse norms (McRae and Buchanan). However, evaluating performance by measuring correlation can be misleading for small and dense norms. We show this by introducing a baseline that replaces the original Binder feature values with nonsensical numbers and then tries to predict the resulting ranking. Concretely, we assign each feature-value pair the difference in character length between concept and feature as a value (\emph{\mbox{CDiff}}, for \emph{C}haracter \emph{Diff}erence). This baseline is different from the \emph{Rand} baseline, because the resulting numbers are more structured.

\begin{table}
\centering
\caption{Learning irrelevant knowledge by mapping fake norms to embeddings. \strut\\}\label{tab:random}

\begin{tabular}{l cccc}
\toprule
\textbf{Norm} & \textbf{Sys} & \textbf{CDiff} & \textbf{Upper}  \\

\midrule
McRae (F1@10) & 0.25 &  0.01 & 0.01\\
Buchanan (F1@10) & 0.18  & 0.01& 0.01\\ \midrule
Binder ($\rho$) & 0.74  & 0.71 & 0.73\\
\bottomrule
\end{tabular}
\end{table}
The results in Table \ref{tab:random} show that we can learn random knowledge like ranking by character length difference, but only from dense matrices like Binder. The resulting  $\rho=0.73$  is not significantly different from the original system with $\rho=0.74$ ($p>0.1$). This is probably due to the \emph{CDiff}-baseline putting so many features on equal ranks. Similar to the data sparsity in the categorical norms, this lack of information leads to uninterpretable evaluation results, at least when only rank correlation is measured. The evaluation clearly shows that nonsensical rankings with specific properties can look like the norms have a significant information overlap with the embeddings.

\emph{Takeaway: Dense norms filled with non-distinctive, structured nonsensical values like \emph{CDiff} can also be predicted from BERT vectors, because the evaluation metric is unsuitable for some rankings. }

\section{Mapping explains geometric similarity}\label{exp:structure}
We have demonstrated that the results of mapping embeddings to feature norms are sensitive to methodological and data-related limitations, which can overshadow the impact of actual information overlap that the experiments aim to elucidate. Subsequently, we will employ a different evaluation measure illustrating that certain geometric similarities in the source and target matrices are crucial for a successful outcome. We then contend that this structural, rather than property-based, information is directly recoverable through the mapping algorithms.

\subsection{Evaluating geometric similarity}\label{sec:structeval}
To provide a structure-based measure, we present the results of neighborhood accuracy for the different experiments. Neighborhood accuracy indicates how many of its direct neighbors from the feature norms are retained in the predicted matrix. The direct neighbors are computed using cosine similarity.

\begin{table}
    \centering

        \caption{Neighborhood Accuracy (NA$@$10) for the original mapping (Sys) and the baselines, including upper bounds (Up). \strut\\}
    \begin{tabular}{l cc cc cc  cc}
        \toprule
    & \multicolumn{2}{c}{\textbf{Sys}} &  \multicolumn{2}{c}{\textbf{Rand}} &  \multicolumn{2}{c}{\textbf{Shuffle}}   &\multicolumn{2}{c}{\textbf{CDiff}}  \\
         Norm & NA & Up & NA & Up & NA & Up & NA & Up \\
        \midrule
        McRae & 0.37 & 0.49 & 0.19 & 0.22 & 0.19 &0.23 &  0.25 & 0.93 \\
        Buchanan & 0.18 & 0.28 & 0.02 & 0.04 &0.02 &0.07 & 0.06 & 0.86 \\  \midrule
        Binder & 0.56 & 0.92 & 0.19 & 0.45 & 0.19 &0.47 & 0.30 & 0.95 \\
        \bottomrule
    \end{tabular}

    \label{tab:na}
\end{table}
Table \ref{tab:na} shows the results for the original system and the baselines we provided in the previous sections. We evaluate the 10 most similar neighbors (NA@10) and give the upper bound for each experiment.
The results show many effects we would have expected for property-based evaluation if that was a genuine explainability method:
\begin{itemize}
\item The \emph{upper bounds} are much higher for the original experiment and particularly so in the \emph{CDiff} baseline. The perfect information overlap here shows in high evaluation results compared to predictions from embeddings, particularly for the structured CDiff matrix.
\item The \emph{random baselines} result in equivalent (low) figures, independently of the way random features are selected. For geometric similarity, the sparseness of the vectors does not make a difference. 
\item \emph{Nonsensical scores} seem not predictable anymore. For Binder, the \emph{CDiff} baseline is still better than the random baselines, but it performs much worse than the original prediction experiment, and is much lower than the upper bound.
\end{itemize}
Overall, the evaluation measure equally applies to sparse and continuous norms. It shows that the information overlap is detectable by mapping embeddings to norms, but the information shared is geometric similarity, not individual features. Consequently, the explanatory power of mapping embeddings to norms lies in predicting similarities between concepts, not in the emergence of actual features or feature-based knowledge.



\subsection{Why geometric similarity is not feature similarity}
 One could argue that similarities in vector space already give rise to a property-based interpretation: In a knowledge base like feature norms, properties are, of course, the source of vector similarities. If we can now say that the same vector similarities are found in the embeddings and determine the accuracy of mapping embeddings to feature norms, we could argue that those properties emerged, because we saw features that span a similar vector space.
While the discovery of geometric parallelism is undeniable, we would be cautious about mistaking correlation for explanation here. If two pairs of vectors have comparable cosine similarity, this is independent of the features that cause this geometric proximity. Accordingly, previous experiments have shown that specific feature knowledge is not derivable overall. 

The primary reason for the unexpected results in feature-based evaluation is the inherent sparsity of categorical norms, which is almost inevitable. If features were not sparse, they would not effectively distinguish the concepts from one another. Some properties are even too rare to be learned at all: 67\% of the McRae features and 37\% of the Buchanan features occur with just one concept (80\% and 52\%, respectively, occur once or twice). In this setup, most concepts will have no feature overlap with each other; thus, the prediction of the nearest neighbors mostly amounts to predicting concept pairs with a similarity greater than zero.

Arguably, geometric similarities between embeddings and norms have a common cause. In particular, taxonomic similarities are represented in distributional embeddings, in general, \citep{pado-lapata-2007-dependency} and BERT embeddings, in particular, \citep{abstractionllm:lrec2024}. They give rise to many similarities in properties because two concepts of the same kind (e.g., two animals) have naturally more in common than two of entirely different types (e.g., an animal and a building). The parallels thus look more like pure correlation and less like explanation.

\emph{Takeaway: Mapping embeddings to feature norms can capture the geometric similarity of the two matrices, but not the specific properties that accounted for it in the target matrix. The sparse norms are especially unsuitable for explaining emergent features.}

\section{Discussion and limitations}\label{discussion}
We presented multiple experiments to illuminate the explanatory capacity of property inference methods. While we aimed to provide comprehensive evidence, our results were limited by the datasets and conclusions we could draw from our experiments.
First, it is vital to assess the explainability method for its trustworthiness. While information overlap is one factor that influences the performance of predicting features from embeddings, we have shown that the geometric peculiarities of the feature norms play a decisive role. Thus, a particularly good or bad performance cannot solely be ascribed to information overlap or the lack thereof. In particular, this makes it impossible to interpret the results in a way that compares the performance for two different norms or states, or to determine which type of feature is better represented than another. Our setup is limited in several respects, which do not compromise our central claims:
\paragraph{Feature norms:} We used three commonly used feature norms for our experiments, two categorical and one continuous dataset. The properties of categorical norms apply to other norms, too, because categorical norms are inherently sparse. We did, however, not simulate experiments that evaluate different degrees of sparsity or different geometric shapes. Relating those metrics to explanatory potential is an important subject of future work.
\paragraph{Embeddings:} Our experiments use layer 0 of BERT embeddings. We have chosen BERT because LLM explainability is a pressing, unresolved issue, and many recent papers on explainability use BERT as their basis. As shown by the upper bounds, our core assertions are more related to the limits of the feature norms and the methods than to the source embeddings. Furthermore, our goal was to demonstrate the influence of methodological limitations on feature prediction as an explainability method, which extends to arbitrary embeddings.
\paragraph{Method:} In our experiments, we described PLSR and FFNNs, the most common property inference methods in previous work. Other algorithms for property inference do exist. As long as those methods are parametrized predictive models, we firmly believe that the theoretical considerations relating to method equivalence also apply to them. The restrictions we present are primarily associated with the nature of the datasets, and this will pose the same challenge for other methods.
\paragraph{Description of explainability:} We have shown that prediction methods can explain geometric similarities between embeddings and property knowledge bases. We do not give a complete account of what else could be explained to what degree, and under which circumstances emerging property knowledge could be analyzed with inference methods. This will be subject to a range of future analyses.

\section{Conclusion}  \label{conclusion}
We provided a detailed analysis of the explanatory capacity of property inference, a frequently used method to explain emerging feature-based knowledge in word embeddings. Our experiments show that the results of property inference cannot be directly taken as faithful explanations, because high accuracy in feature prediction does not indicate high information overlap between the embeddings and the feature knowledge base. Instead, methodological upper bounds and geometric similarities determine the results.

In more detail, we have shown that hyperparameter tuning is vital and often not done correctly, and that the most popular methods (PLSR and FFNN) are essentially equivalent. We then used multiple ablation experiments that prove that the performance of property inference is primarily determined by the methodology's upper bound, which is, again, based on the particular shape of the input data. Most of the feature norms are not well-predictable even with perfect information overlap, and, at the same time, knowledge sources with equal information overlap (e.g., two different types of random features) seem to perform differently, depending on the sparsity of the matrix.

We also showed that random features and random rankings can be predicted to some degree (the former for sparse norms and the latter for categorical norms).  Using a different evaluation measure that targets geometric similarity, we could show what actually \emph{is} explainable by property inference methods, which amounts to the similarity of the vector spaces spanned. This must be assessed with appropriate measures and is not the same as property knowledge, but similar geometric structures are effects with a common cause.

Our results show that some intuitive interpretations of this explainability approach are theoretically and heuristically questionable. A closer inspection of correlation-based interpretability methods, in general, is essential to assess which types of predictions actually reveal knowledge and which side effects are merely suggested but not supported by the theoretical background. In future work, the explanatory power of explanation approaches for embeddings needs to be analyzed in more detail, such as its dependence on sparsity, geometric structures, properties of scores, and general capacity for explaining the emergence of specific knowledge.



\bibliography{abstraction}

\clearpage


\appendix

\section{Details on Hyperparameter Search} \label{app:overfitting}

In Section \ref{sec:method}, we stated that most previous approaches use parameter settings for PLSR and FFNN that overfit on the small amounts of input data. We provided the optimal parameters for each setting and the graphs showing training and test performance for the McRae norms. We show detailed results for each algorithm and norm in the following. For training and test values in 10-fold cross-validation, the 10 runs are averaged.  The graph for McRae is repeated for the sake of completeness.

\begin{figure}[H]
    \includegraphics[scale=0.48]{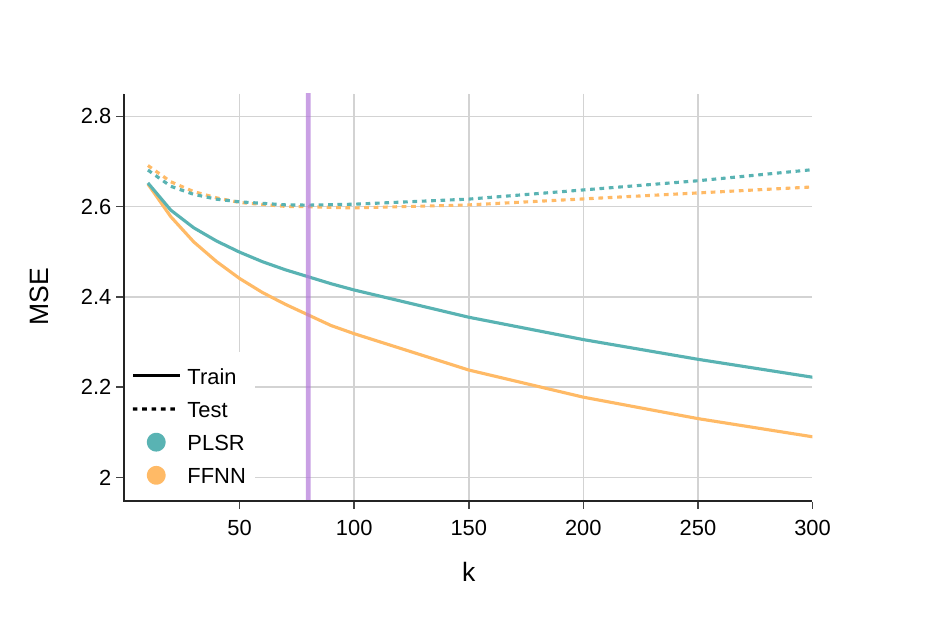}\caption{Mean Squared Error (MSE) for System (Original experiment) on Buchanan using PLSR with number of components / FFNN with hidden layer sizes.}
    \includegraphics[scale=0.48]{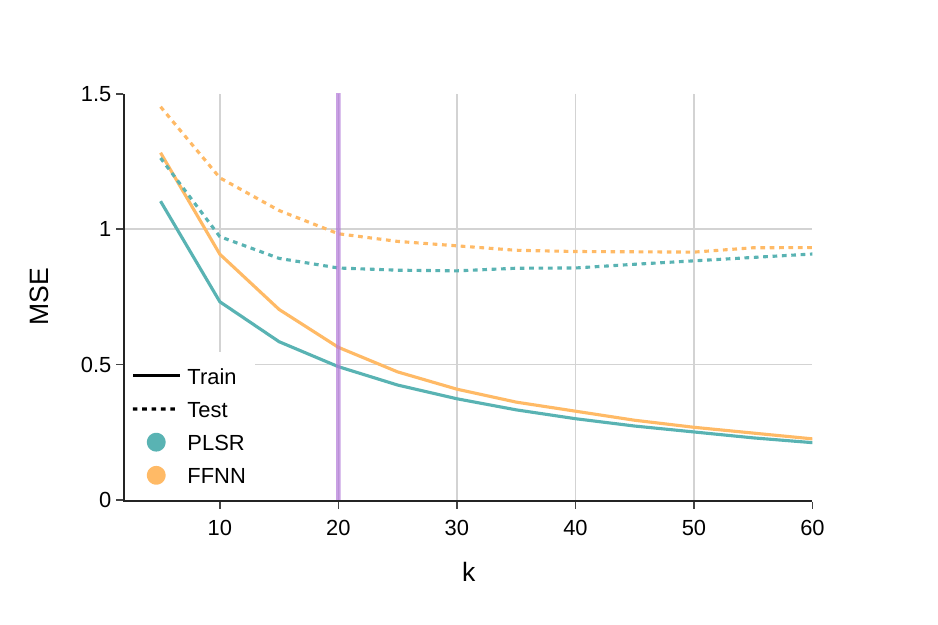}\caption{Mean Squared Error (MSE) for System (Original experiment) on Binder using PLSR with number of components / FFNN with hidden layer sizes.}
        \includegraphics[scale=0.48]{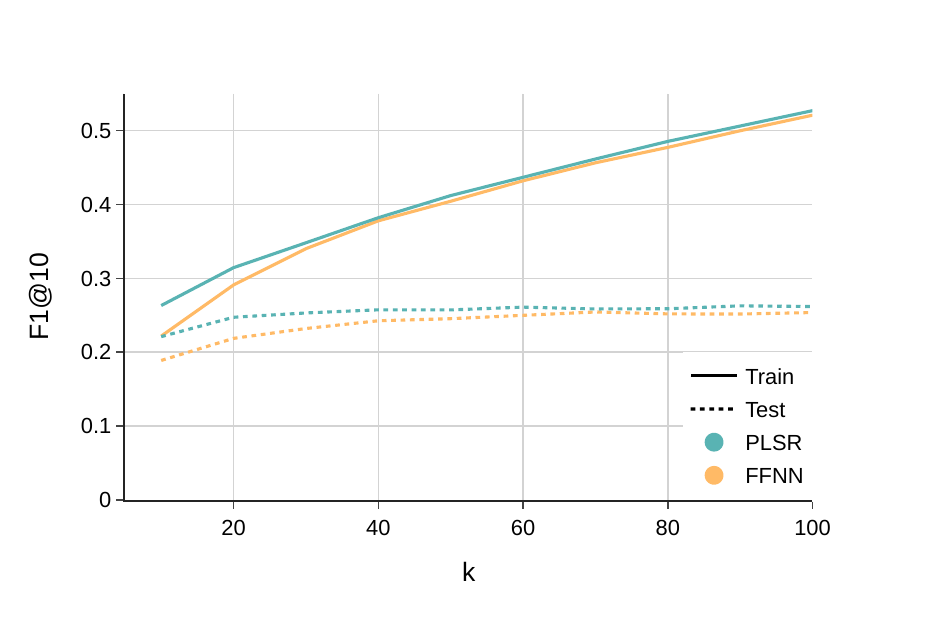}\caption{F1@10 for System (Original experiment) on McRae using PLSR with number of components / FFNN with hidden layer sizes.}
\end{figure}
\begin{figure}[H]

    \includegraphics[scale=0.48]{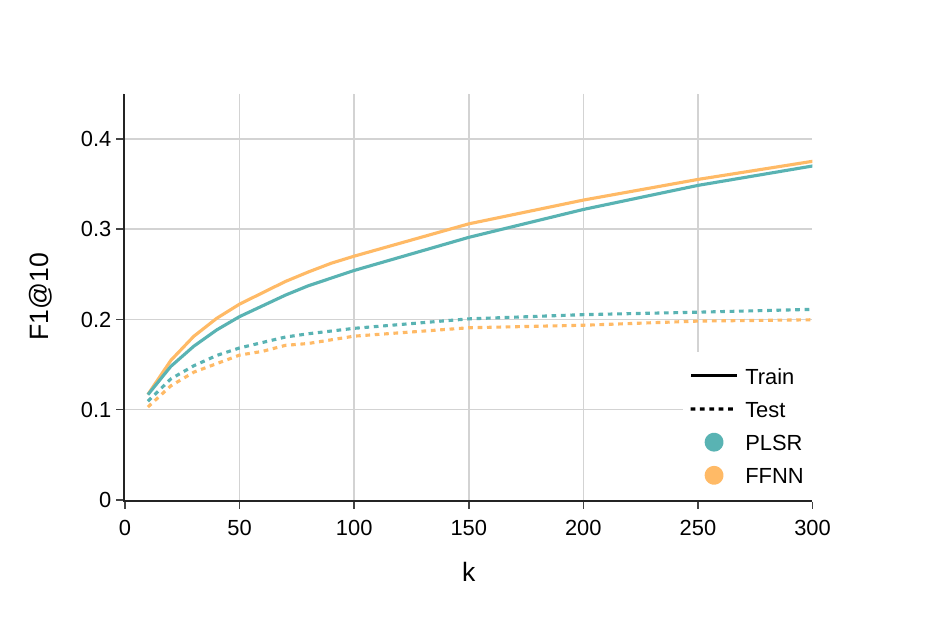}\caption{F1@10 for System (Original experiment) on Buchanan using PLSR with number of components / FFNN with hidden layer sizes.}
    \label{fig:overfitting-buchanan-f1}

    \includegraphics[scale=0.48]{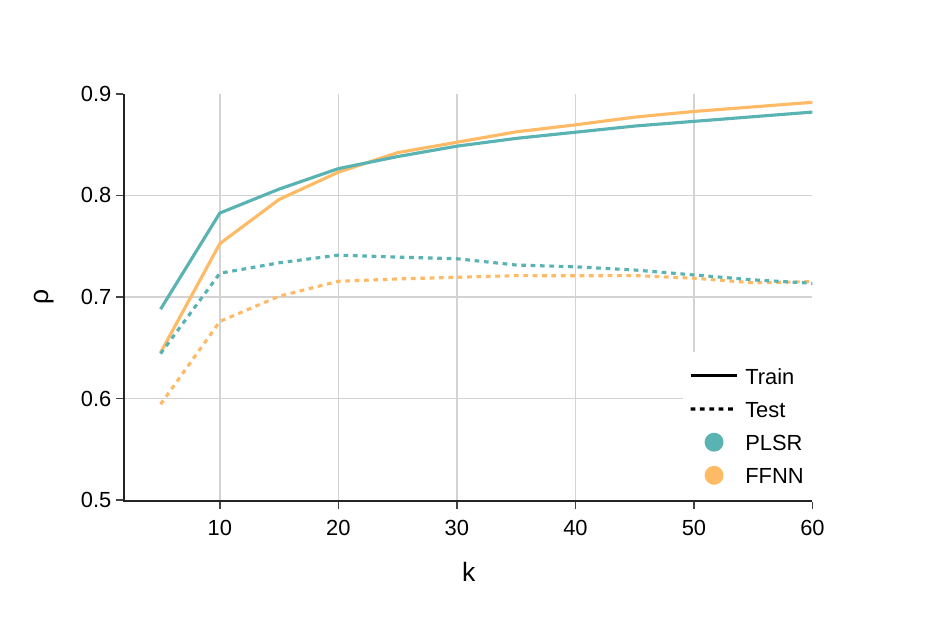}\caption{$\rho$ (Spearman's correlation) on Binder using PLSR with number of components / FFNN with hidden layer sizes.}
    \label{fig:overfitting-binder-rho}
\end{figure}

\begin{table}[H]
  \centering
  
  \caption{Overfitting. F1@10, MSE, Spearman's $\rho$ and Neighbors Accuracy@10 on PLSR and FFNN with different number of components/hidden layer size.\strut\\}
  \begin{tabular}{l cc cc}
      \toprule
      \textit{McRae} &  \multicolumn{2}{c}{PLSR} & \multicolumn{2}{c}{FFNN} \\
      \cmidrule(lr){2-3} \cmidrule(lr){4-5}
      Metric & 25 & 300 & 25 & 300  \\
      \midrule
      Train F1 & 0.33 & 0.79 & 0.32 & 0.73 \\
      Test F1 & 0.25 & 0.24 & 0.23 & 0.26 \\ \midrule
      Train MSE & 0.59 & 0.19 & 0.62 & 0.30 \\
      Test MSE & 0.67 & 0.93 & 0.68 & 0.75 \\
      \toprule
      \textit{Buchanan} &  \multicolumn{2}{c}{PLSR} & \multicolumn{2}{c}{FFNN} \\
      \cmidrule(lr){2-3} \cmidrule(lr){4-5}
      Metric & 80 & 300 & 80 & 300  \\
      \midrule
      Train F1 & 0.24 & 0.37 & 0.25 & 0.38 \\
      Test F1 & 0.18 & 0.21 & 0.17 & 0.20 \\ \midrule
      Train MSE & 2.44 & 2.22 & 2.36 & 2.09 \\
      Test MSE & 2.60 & 2.68 & 2.60 & 2.64 \\
      \toprule
      \textit{Binder} &  \multicolumn{2}{c}{PLSR} & \multicolumn{2}{c}{FFNN} \\
      \cmidrule(lr){2-3} \cmidrule(lr){4-5}
      Metric & 20 & 50 & 20 & 50  \\
      \midrule
      Train $\rho$ & 0.83 & 0.87 & 0.82 & 0.88 \\
      Test $\rho$ & 0.74 & 0.72 & 0.71 & 0.72 \\ \midrule
      Train MSE & 0.49 & 0.25 & 0.56 & 0.27 \\
      Test MSE & 0.86 & 0.88 & 0.99 & 0.91 \\ \midrule
      Train NA & 0.49 & 0.66 & 0.56 & 0.71 \\
      Test NA & 0.56 & 0.54 & 0.54 & 0.54 \\
      \bottomrule
  \end{tabular}

  \label{tab:overfitting-table}
\end{table}
\clearpage

\section{Extended Evaluation Results}
For space reasons and because they are equivalent, we only reported the results for PLSR in the main part of the paper. For reference, we also report the results for FFNNs here.

\begin{table}[H]
    \centering
    \caption{Mapping results for shuffled features and a random baseline using FFNN (shown in Sec.\ref{sec:upper}) and \ref{sec:shuffle}.\strut\\ }

    \begin{tabular}{l cc cc cc}
    \toprule
    \textbf{Norm} & \textbf{Original} & \textbf{Shuffle} & \textbf{Random}  \\
    \midrule
    McRae (F1)&  0.23 & 0.11 & 0.01 \\
    Buchanan (F1) & 0.17 & 0.06 & 0.01 \\ \midrule
    Binder ($\rho$) & 0.71 & 0.33 & 0.00 \\
    \bottomrule
    \end{tabular}
    \label{tab:shuffle-ffnn}
\end{table}

\begin{table}[H]
    \centering
        \caption{Neighborhood Accuracy (NA$@$10) for the original mapping and the baselines using FFNN (shown in Sec.\ref{sec:structeval}).\strut\\ }

    \begin{tabular}{l cc cc cc}
        \toprule
    Norm & \textbf{Original} & \textbf{Random} & \textbf{CDiff} \\
        \midrule
        McRae & 0.36 & 0.18 & 0.33 \\
        Buchanan & 0.17 & 0.02 & 0.06 \\  \midrule
        Binder & 0.54 & 0.19 & 0.31 \\
        \bottomrule
    \end{tabular}
    \label{tab:na-ffnn}
\end{table}

\end{document}